\documentclass[10pt,twocolumn,letterpaper]{article}

\usepackage{iccv}
\usepackage{times}
\usepackage{epsfig}
\usepackage{graphicx}
\usepackage{amsmath}
\usepackage{amssymb}
\usepackage{algorithm}    
\usepackage{algorithmic}
\usepackage{bm}
\usepackage{color}
\usepackage{diagbox}

\usepackage[pagebackref=true,breaklinks=true,letterpaper=true,colorlinks,bookmarks=false]{hyperref}
\usepackage[hyphenbreaks]{breakurl}

\iccvfinalcopy 

\def\iccvPaperID{4562} 

\ificcvfinal\fi
\begin{document}

\title{Adversarial Learning with Margin-based Triplet Embedding Regularization}

\author{Yaoyao Zhong, Weihong Deng\thanks{Corresponding Author}\\
Beijing University of Posts and Telecommunications\\
{\tt\small \{zhongyaoyao, whdeng\}@bupt.edu.cn}
}

\maketitle

\begin{abstract}
  The Deep neural networks (DNNs) have achieved great success on a variety of computer vision tasks, however, they are highly vulnerable to adversarial attacks. To address this problem, we propose to improve the local smoothness of the representation space, by integrating a margin-based triplet embedding regularization term into the classification objective, so that the obtained model learns to resist adversarial examples. The regularization term consists of two steps optimizations which find potential perturbations and punish them by a large margin in an iterative way. Experimental results on MNIST, CASIA-WebFace, VGGFace2 and MS-Celeb-1M reveal that our approach increases the robustness of the network against both feature and label adversarial attacks in simple object classification and deep face recognition.The code is available at \url{https://github.com/zhongyy/Adversarial_MTER}
\end{abstract}

\section{Introduction}
The Deep neural networks (DNNs) have achieved great success~\cite{krizhevsky2012imagenet,simonyan2014very,he2016deep,hu2017squeeze}, significantly improving the development of a variety of challenging applications such as deep face recognition~\cite{Chen2014Deep,Schroff2015FaceNet,Wen2016A,Liu2017SphereFace,Wang2018CosFace,deng2018arcface,DBLP1,DBLP2} and automatic driving~\cite{Bojarski16,Codevilla18}. 

However, contradictions between the vulnerability of DNNs and the demand of security have become increasingly obvious. On one hand, DNNs are vulnerable. Previous works have discovered, with elaborate strategies, DNNs can be easily fooled by test images with imperceptible noise~\cite{Szegedy14}. This type of images is named as adversarial examples. Moreover, adversarial examples are transferable in different models~\cite{Papernot17,dong2018boosting}, which means black-box attacks can be launched without knowing the details of target models (\eg architectures, parameters and defense methods). On the other hand, the demand of security arises in safety crucial domains driven by DNNs. Adversarial examples can attack physical-world DNNs applications~\cite{Kurakin17}. For instance, DNNs in an automatic vehicle system can be confused by carefully manipulated road signs~\cite{Eykholt2018}, and DNNs in a face recognition system are susceptible to feature level adversarial attacks~\cite{Sabour16,Rozsa2017,Song18}. 

The existence of adversarial examples has given birth to a variety of researches on adversarial defenses. One straightforward defense strategy is to increase the robustness of the model by injecting adversarial examples in the training process ~\cite{Szegedy14,Kurakin17atscale,Tramer18}, which is essentially a regularization of the training data augmentation. This strategy is effective in close-set classification like object classification, while may not suitable in open-set settings like deep face recognition where training categories could be in million level. Another strategy is to detect adversarial examples at inference time~\cite{Metzen17,Xu18,Song18}. This strategy is appropriate for both open-set and close-set classification settings, while it can be easily broken in the white-box setting where the specific defense method is known~\cite{Carlini17}. 

\begin{figure}[htbp]
	\center
	\includegraphics[width=1\linewidth]{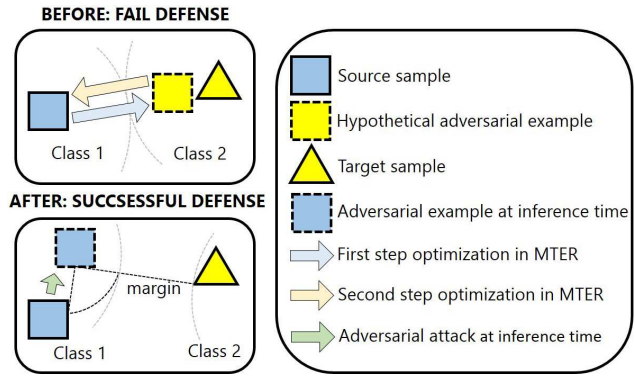}
	\caption{Schematic illustration of the defense before MTER training (top) versus after training (bottom). Arrows indicate the gradients arising from the optimization of the cost function. The same color represents the same predict class.}
	\label{fig:ach}
\end{figure}

In this paper, we propose a margin-based triplet embedding regularization (MTER) method to train DNNs with robustness. Our intuition is that by training a model to improve the local smoothness of embedding space with fewer singular points, it will be more resistant to adversarial examples. The regularization term consists of two steps optimizations in an iterative way which first find potential perturbations in the embedding space, and then punish them by a large margin. A schematic illustration of the defense before MTER versus after MTER is shown in Figure~\ref{fig:ach}. Specifically, in the embedding space, a potential attack is generated from a source to a target. We improve the robustness by encouraging the hypothetical attacks to gradually approach the source class, meanwhile move far away from all the other target classes. The result of an embedding space visualization experiment is shown in Figure~\ref{fig:toy}. In the optimization, the large margin is not trivial, which strictly ensures the inter-class distance and the intra-class smoothness in the embedding space. Our contributions are as follows: 

1. We propose to improve the robustness of DNNs by smoothing the embedding space, which is appropriate for DNNs trained in both open-set and close-set classification settings.

2. We introduce the large margin into adversarial learning, which further guarantees the inter-class distance and the intra-class smoothness in the embedding space, therefore improves the robustness of DNNs.

3. Experimental results on MNIST~\cite{LeCun98}, CASIA-WebFace~\cite{Yi2014CASIA}, VGGFace2~\cite{Cao18} and MS-Celeb-1M~\cite{Guo16MS} demonstrate the effectiveness of our methods in simple object classification and deep face recognition.

\section{Related work}
\subsection{Adversarial attacks}
Szegedy \etal~\cite{Szegedy14} first find that they can cause DNNs to misclassify images by a certain hardly perceptible perturbation which is generated using a box-constrained L-BFGS method. Compared with the L-BFGS attack~\cite{Szegedy14}, Goodfellow \etal~\cite{Goodfellow15} propose a more time-saving and practical method "fast`` method (FGSM) to generate adversarial examples by performing one-step gradient update along the direction of the sign of gradient at each pixel.

Kurakin \etal~\cite{Kurakin17} introduce a straightforward method called "basic iterative`` method (BIM), to extend the "fast`` method (FGSM)~\cite{Goodfellow15} by applying it multiple times with small step size and clip pixel values after each step to ensure the $L_\infty$ constraint. Moreover, to generate adversarial examples of a specific desired target class, Kurakin \etal~\cite{Kurakin17} introduce the iterative least likely method. Iterative methods could attacks DNNs with higher rate compared with the fast method in the same constraint level~\cite{Kurakin17}. Similarly, another iterative attack method proposed by Moosavi-Dezfooli \etal~\cite{Dezfooli16}, called Deepfool, is also reliable and efficient. With linear approximation, Deepfool try to generate the minimal perturbation in each step by moving towards the lineared decision boundary~\cite{Dezfooli16}. Based on Deepfool~\cite{Dezfooli16}, Moosavi-Dezfooli \etal~\cite{Dezfooli17} propose image-agnostic adversarial attacks, which could fool DNNs with a universal perturbation on images with high probability.

Apart from the generation of adversarial examples, there are also works focus on the transferability of adversarial examples~\cite{Papernot17,dong2018boosting,LeiWu18}, adversarial examples in physical world~\cite{Kurakin17,Eykholt2018}, and in specific tasks such as face recognition system~\cite{Sharif16,Goswami18}.

\subsection{Defense methods}

The defense methods can be classified into two categories, one is to improve the robustness of DNNs, the other one is to detect adversarial examples at inference time~\cite{Metzen17,Xu18,Song18}. We mainly discuss the former type, which is more related to our work. 

Network distillation~\cite{Jimmy14,Hinton15} is originally proposed to reduce the model size. Papernot \etal~\cite{Papernot16} introduce distillation as a defense method to improve the robustness by feeding back the class probability to train the original model. 

Adversarial training could provide regularization to DNNs~\cite{Goodfellow15}. Goodfellow \etal~\cite{Goodfellow15} first propose adversarial training which could increase the robustness by injecting adversarial examples in the training process. Then adversarial training is applied and analyzed in large training dataset ImageNet~\cite{Kurakin17atscale}. 

Although the success of adversarial training on white-box defenses, the defense against black-box attacks is still a problem, due to the transferability of adversarial examples. To deal with the transferred black-box attacks, Tramer \etal~\cite{Tramer18} introduce ensemble adversarial training technique transferring one-step adversarial examples from other training models, while Na \etal~\cite{Na18} propose cascade adversarial trained transferring iterative attacks from already trained model.

Around the same time, Dong \etal~\cite{Dong17} and Na \etal~\cite{Na18} minimize both the cross-entropy loss and the distance of original and adversarial embedding to improve the vanilla adversarial training. They are the most related work to ours. However, our method mainly differs from them in two aspects: (1) Our MTER method is a straightforward and thorough feature adversary which will not be limited by number of training categories, therefore is also appropriate for open-set classification. (2) We introduce the large margin into adversarial learning, which guarantees not only the intra-class smoothness but also the inter-class distance in the embedding space.

\section{Margin-based Regularization}
Our purpose is training a DNN to have smooth representations with fewer singular points. Therefore we consider a regularization term which exploits the vulnerability and further fix them in an iterative way. 

\begin{algorithm}[htbp]
	\begin{algorithmic}
		\caption{Margin-based triplet embedding regularization (MTER)}
		\label{alg:1}
		\renewcommand{\algorithmicrequire}{\textbf{Input:}}
		\REQUIRE
		\STATE Training set $D = \{ {x^{(i)}},{y^{(i)}}\in \{ 1,2,...,C\}\}$, model parameter $\theta$ and hyperparameter margin $m$, mini-batch size $K$.
		\renewcommand{\algorithmicrequire}{\textbf{Output:}}
		\REQUIRE
		\STATE The final model parameter $\theta$.
		\renewcommand{\algorithmicrequire}{\textbf{Initialization at the beginning of an epoch:}}
		\REQUIRE
		\STATE // Constructing the source images queue $Qs$ and the $Qt$. 
		\STATE $Qs$=$Qt$=\{\}. 
		\STATE Random select $\frac{C}{2}$ categories in $\{ 1,2,...,C\}$ denoted as source $S$, the complementary set is target $T$.
		\STATE $Qs$.append($\{x^i|y^i \in S\}$);
		$Qt$.append($\{x^i|y^i \in T\}$).
		\STATE shuffle $Qs$ and $Qt$.
		\renewcommand{\algorithmicensure}{\textbf{Optimization in an epoch:}}
		\ENSURE
		\WHILE{$Qt$ is not empty and $Qs$ is not empty}  
		\STATE Take out a mini-batch $Bs$ and $Bt$ with $\frac{K}{2}$ samples respectively in $Qs$ and in $Qt$.
		\STATE $\Delta {x^{(s,t)}}\leftarrow$ Calculate perturbations~\eqref{a:3} in an iterative way~\eqref{a:5} on the batch $Bs$ and $Bt$, based on the current model $\theta$.
		\STATE $\theta \leftarrow \nabla(L_{ori}~\eqref{a:2}+R_{M\!T\!E\!R}~\eqref{a:10}) $ on batch $Bs$, $Bt$ and $\Delta {x^{(s,t)}}$.
		\ENDWHILE 
	\end{algorithmic}  
\end{algorithm} 

\subsection{Exploitation the Vulnerability}
First we consider the vulnerability exploitation. We start from some notations. Let $D = \{ {x^{(i)}},{y^{(i)}}\}$ denote a labeled dataset where ${x^{(i)}}$ and ${y^{(i)}} \in \{ 1,2,...,C\}$ respectively denote an input image and the corresponding label. A DNN can be formulated in a chain\begin{equation}{F_\theta }^{(n)}(x) = {f^{(n)}}(...({f^{(2)}}({f^{(1)}}(x))),\end{equation} parameterized by $\theta$. The network is originally trained on a dataset $D$ by cross entropy
\begin{small}\begin{equation}\label{a:2}L_{ori} \!=\! \mathop {\arg \min }\limits_\theta  H({F_\theta }^{(n)}({x^{(i)}}),{y^{(i)}}) \!= \! - \frac{1}{K}\sum\limits_{i = 1}^K {\log p({y^{(i)}}|{x^{(i)}})}, \end{equation}\end{small}where $H$ giving the sum of the cross entropies between the predictions ${F_\theta }^{(n)}({x^{(i)}})$ and the labels ${y^{(i)}}$. 

Given a trained DNN, a source image and a target image, denoted as $\{ {x^{(s)}},{x^{(t)}}\}$, where ${y^{(s)}} \ne {y^{(t)}}$, we could find small perturbations  $\Delta {x^{(s,t)}}$ to the source image  ${{x^{({\rm{s}})}}}$ that produce an internal representation that is remarkably similar to that of the target image ${{x^{({\rm{t}})}}}$~\cite{Sabour16}. The vulnerability exploitation in embedding space can be described as:
\begin{small}\begin{equation}
\label{a:3}
\Delta {x^{(s,t)}} = \mathop {\arg \min }\limits_{\Delta {x^{(s,t)}}} \left\| {E({x^{(s)}} + \Delta {x^{(s,t)}}) - E({x^{(t)}})} \right\|_2^2,
\end{equation}\end{small}
subject to \begin{equation}
{\left\| {\Delta {x^{(s,t)}}} \right\|_\infty } < \varepsilon.
\end{equation}
$E(x)$ is the deep representation in the embedding space, which is normalized to unit length from ${F_\theta }^{(n - 1)}(x)$. ${F_\theta }^{(n - 1)}(x)$ is the function from the image $x$ to its representation at the $n-1$ layer. $\varepsilon$ limits the maximum deviation of the perturbation. 

For computational efficiency, we adopt the direction defined by the gradient of the metric loss function and form adversarial perturbations in an iterative way, referred to as iterative feature target gradient sign method (IFTGSM):
\begin{small}\begin{equation}\label{a:5}\begin{gathered}
\Delta x_0^{(s,t)} = 0, \hfill \\
{x^{({\text{s}})}} + \Delta x_{N + 1}^{(s,t)} = C{_{{x^{({\text{s}})}},\varepsilon }}({x^{({\text{s}})}} + \Delta x_N^{(s,t)} +  \hfill \\
\qquad\quad sign(\nabla_{{x^{(s)}} + \Delta x_N^{(s,t)}}\left\| {E({x^{(s)}} + \Delta x_N^{(s,t)}) - E({x^{(t)}})} \right\|_2^2)), \hfill \\
\end{gathered}\end{equation}\end{small}where\begin{equation}\label{a:6} C{_{x,\varepsilon }}(x') = \min (255,x + \varepsilon ,\max (0,x - \varepsilon ,x')),\end{equation} the iteration is chosen heuristically $\min (\varepsilon+4,1.25\varepsilon)$. $\Delta {x^{(s,t)}}$ can also be generated using a fast method, referred to as fast feature target gradient sign method (FFTGSM). We formulate it as follows: \begin{equation}\label{a:7}\Delta {x^{(s,t)}} = {\text{  }}\varepsilon sign({\nabla _{{x^{(s)}}}}\left\| {E({x^{(s)}}) - E({x^{(t)}})} \right\|_2^2)). 
\end{equation} We will use this fast attack method in the experiment on face recognition in Section~\ref{section:face}.

\subsection{Fix the Vulnerability}
Our target is to improve the robustness of DNNs without modifying their architectures. The aforementioned vulnerability attacks a DNN by finding singular points in the internal representation space of a DNN. Considering the existence of the vulnerability, we find it is possible that we smooth the embedding space by jointly optimizing the original cross entropy and a large-margin based triplet distance constraint as a regularization term. 
  
With a source and a target image $\{ {x^{(s)}},{x^{(t)}}\}$, consider a triplet $t:=\{ E({x^{(s)}}), E({x^{(s)}+\Delta {x^{(s,t)}}}),E({x^{(t)}})\}$, where $\Delta {x^{(s,t)}}$ is the aforementioned perturbation. Ideally, for all triplets $t$ which are generated in the training set, we would like the following constraint to be satisfied:
\begin{small}\begin{equation}\left\| {E({x^{({\text{s}})}}\!\! +\!\! \Delta {x^{(s,t)}})\!\! -\!\! E({x^{(s)}})} \right\|_2^2\!\! <\!\! \left\| {E({x^{(s)}}\!\! +\!\! \Delta {x^{(s,t)}})\!\! -\!\! E({x^{(t)}})} \right\|_2^2.\end{equation}\end{small}However, due to the first step optimization in objective~\eqref{a:3}, the actual situation at a certain moment in the training process may be:
\begin{small}\begin{equation}\left\| {E({x^{({\text{s}})}}\!\! +\!\! \Delta {x^{(s,t)}})\!\! -\!\! E({x^{(s)}})} \right\|_2^2\!\! >\!\! \left\| {E({x^{(s)}}\!\! +\!\! \Delta {x^{(s,t)}})\!\! -\!\! E({x^{(t)}})} \right\|_2^2.\end{equation}\end{small}

Therefore, the vulnerability exploitation and fixing together constitute two optimization steps in the adversarial learning, which strive to attack each other but also together improve the robustness of DNNs gradually. Formally, we define the margin-based triplet embedding regularization (MTER) as follows:
\begin{small}
\begin{equation}
\label{a:10}
\begin{gathered}
R_{M\!T\!E\!R}\!\!=\!\!\frac{1}{K}\!\!\!\!\!\!\sum\limits_{{y^{(s)}} \ne {y^{(t)}}}\!\!\!\! {\max (0} ,m \!+\! \left\| {E({x^{({\text{s}})}} \!+\! \Delta {x^{(s,t)}}) \!-\! E({x^{(s)}})} \right\|_2^2 \hfill \\
\qquad\qquad\qquad\qquad\qquad\quad\;- \left\| {E({x^{(s)}} \!+\! \Delta {x^{(s,t)}}) \!-\! E({x^{(t)}})} \right\|_2^2), \hfill \\ 
\end{gathered} 
\end{equation}
\end{small}where the $\Delta {x^{(s,t)}}$ is obtained and upgraded by objective~\eqref{a:3}, and the parameter $m$ is the margin. In practice, we apply the vulnerability exploitation and fixing in an iterative way, which is precisely described in Algorithm~\ref{alg:1}. Parameter $m$ controls that the similarity between the source image and the perturbed image should be much higher than that between the perturbed image and the target image. $m$ is chosen based on the training dataset and the model capacity. We will discuss the parameter $m$ in the following ablation study in Section~\ref{sec:m}.

\section{Experiment}
\subsection{Experiment on Simple Image Classification}
In this section, we first analyze the effect of the margin-based triplet embedding regularization (MTER) method on MNIST~\cite{LeCun98}, a simple image classification task. We train ResNet~\cite{he2016deep} models using original training loss functions, adversarial training, and our MTER method, respectively. We test different models assuming that the adversary knows the classification algorithm, model architecture and parameters, because the reliability of a model could be demonstrated if a model is robust in the white-box setting.

We first give a brief description of the adversarial training method~\cite{Kurakin17atscale} and attack methods FGSM~\cite{Goodfellow15}, BIM~\cite{Kurakin17}, FTGSM~\cite{Kurakin17}, ITGSM~\cite{Kurakin17} which we will test and compare with our method. 

Fast gradient sign method (FGSM)~\cite{Goodfellow15} generates adversarial examples by perturbing inputs in a manner that increases the sign of the gradients of the original loss function \wrt the input image $x^{(i)}$:
\begin{equation}
x_{adv}^{(i)} = {x^{(i)}} + \varepsilon sign({\nabla _{{x^{(i)}}}}H({F_\theta }^{(n)}({x^{(i)}}),{y^{(i)}})),
\end{equation}where $H$ giving the cross entropy between the predictions ${F_\theta }^{(n)}({x^{(i)}})$ and the labels ${y^{(i)}}$, $\varepsilon$ limits the maximum deviation of the perturbation. 

Basic iterative method (BIM)~\cite{Kurakin17} is a modification of the FGSM~\cite{Goodfellow15} by applying it multiple times:
\begin{equation}
\begin{gathered}
x_{adv,0}^{(i)} = {x^{(i)}},x_{adv,N + 1}^{(i)} =  \hfill \\
{C_{{x^{(i)}},\varepsilon }}(x_{adv,N}^{(i)} + \alpha sign({\nabla _{x_{adv,N}^{(i)}}}H({F_\theta }^{(n)}(x_{adv,N}^{(i)}),{y^{(i)}}))), \hfill \\ 
\end{gathered}
\end{equation}
where $\alpha = 1$ is used, $C_{{x^{(i)}},\varepsilon }$ is referred to as equation~\eqref{a:6}, and number of iterations is $\min (\varepsilon  + 4,1.25\varepsilon )$. 

Compared with BIM~\cite{Kurakin17}, iterative target gradient sign method (ITGSM)~\cite{Kurakin17} leads the model to misclassify an image as another target category:
\begin{equation}
\label{a:13}
\begin{gathered}
x_{adv,0}^{(i)} = {x^{(i)}},x_{adv,N + 1}^{(i)} =  \hfill \\
{C_{{x^{(i)}},\varepsilon }}(x_{adv,N}^{(i)} - \alpha sign({\nabla _{x_{adv,N}^{(i)}}}H({F_\theta }^{(n)}(x_{adv,N}^{(i)}),y_t^{(i)}))), \hfill \\ 
\end{gathered}
\end{equation}
where $y_t^{(i)}$ is the target label we would like the model to predict, $y_t^{(i)} \ne {y^{(i)}}$. Also, the target attack can be launched in a Fast style, referred to as fast target gradient sign method (FTGSM)~\cite{Kurakin17}: 
\begin{equation}
\label{a:14}
x_{adv}^{(i)} = {x^{(i)}} - \varepsilon sign({\nabla _{{x^{(i)}}}}H({F_\theta }^{(n)}({x^{(i)}}),{y_t^{(i)}})).
\end{equation}

We use ResNet-18~\cite{he2016deep} for training. The models are trained from scratch. There is no data augmentation for both datasets. $\varepsilon=0.3*255$ is applied for MNIST~\cite{LeCun98}. In ITGSM~\cite{Kurakin17} attack, the least likely class is used as the target class. We use two type of original loss functions Softmax and ArcFace~\cite{deng2018arcface}, which is a type of large-margin loss and first used in deep face recognition. The feature scale is set to 10 and angular margin is set to 0.4 in ArcFace~\cite{deng2018arcface}. We improve the robustness of the two type of original loss respectively using adversarial training~\cite{Kurakin17atscale} and our method. The margin $m$ in our MTER method is set to 0.2. We follow the adversarial training implemented in Kurakin \etal~\cite{Kurakin17atscale}, which increases the robustness by replacing half of the mini-batch clean examples with their adversarial examples into the training process. More specifically, we generate adversarial examples using FGSM~\cite{Goodfellow15} perturbations with respect to predicted rather than true labels following works~\cite{Na18,Ross18}, to prevent ''label leaking``~\cite{Kurakin17atscale} where the model tend to learn to classify adversarial examples more accurately than regular examples. The relative weight of adversarial examples in the loss is set to 0.3 following~\cite{Kurakin17atscale}. 

The results are shown in Table~\ref{table:r18_mnist}. As shown in the table, even though adversarial training is done with the predicted label, the label leaking phenomenon~\cite{Kurakin17atscale} still happens. Our MTER method improves the robustness of the original models using different loss functions under FGSM~\cite{Goodfellow15}, BIM~\cite{Kurakin17} and ITGSM~\cite{Kurakin17} attacks. For models trained with Softmax, our method sacrifices a little performance on the clean images. while for models trained with ArcFace~\cite{deng2018arcface}, a large margin loss function, it even improves the accuracy on clean images. Besides, it outperforms adversarial training method under BIM~\cite{Kurakin17} and ITGSM~\cite{Kurakin17} attacks. Even though we did not use these type of adversarial examples for data augmentation in training like adversarial training method, our method could still gain robust improvements under unknown attacks. This indicates that our method can improve the robustness of models on simple image classification like MNIST~\cite{LeCun98}.

\begin{table}
	\renewcommand\arraystretch{1}
	\begin{center}
		\scalebox{0.69}{
			\begin{tabular}{|c|c|c|c|c|}
				\hline
				Method&Clean&FGSM~\cite{Goodfellow15}&BIM~\cite{Kurakin17}&ITGSM~\cite{Kurakin17}\\
				\hline\hline
				Softmax loss&99.6&10.4&0.0&6.0\\
				Adversarial training&99.6&99.9&3.1&47.2\\
				Softmax+MTER (ours) &99.5&96.8&98.7&95.1\\
				\hline\hline
				ArcFace Loss&99.5&28.6&1.7&24.7\\
				Adversarial training&99.5&99.5&30.5&71.3\\
			    ArcFace Loss+MTER (ours) &99.6&96.6&98.0&95.3\\
				\hline
			\end{tabular}
		}
	\end{center}
\vspace{-0.2cm}
	\caption{MNIST~\cite{LeCun98} test results (\%) for Resnet-18~\cite{he2016deep} models ($\varepsilon=0.3*255$ at test time). The higher the accuracy is, the more robust is the target model.}
	\vspace{-1.5ex}
	\label{table:r18_mnist}
\end{table}

\subsection{Embedding Space Visualization}
MNIST~\cite{LeCun98}, the popular and sweet dataset is used for embedding space visualization. We use the ResNet-18~\cite{he2016deep} by changing the original fully connected layer to two fully connected layer and then modifying the embedding dimension to 2. We retrain networks on MNIST~\cite{LeCun98} with Softmax and Softmax combined our MTER method, respectively. Then we use the clean examples of test dataset for visualization.  Besides, for each class, we randomly choose a test sample and generate adversarial examples of it using BIM~\cite{Kurakin17} for visualization. 

\begin{figure}[htbp]
	\center
	\includegraphics[width=1\linewidth]{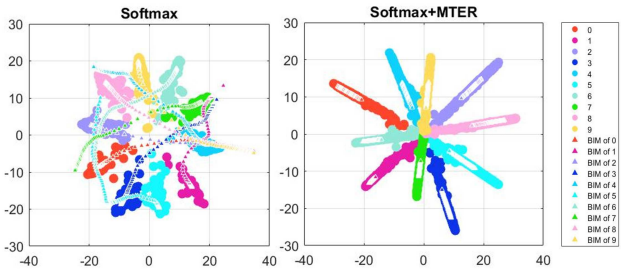}
	\caption{Embedding space visualization on ResNet-18~\cite{he2016deep} which is modified embedding dimension to 2. Models are trained on MNIST~\cite{LeCun98} with Softmax and Softmax combined with MTER method, respectively. The round represents the clean examples of MNIST~\cite{LeCun98} test set. The triangle represents BIM~\cite{Kurakin17} adversarial examples, which we draw from $\varepsilon=0 $ to $\varepsilon=76 (\approx 0.3\times 255)$ for one sample image per each class.}
\vspace{-1.5ex}
	\label{fig:toy}
\end{figure}

The results are shown in Figure~\ref{fig:toy}. The round represents the clean examples of MNIST~\cite{LeCun98} test set. The triangle represents BIM~\cite{Kurakin17} adversarial examples, which we draw from $\varepsilon=0 $ to $\varepsilon=76 (\approx 0.3\times 255)$ for one sample image per each class. We can observe that with our MTER method, the adversarial examples are close to clean examples and distributed in the original class region in the embedding space. The inter-class margin is enlarged and the intra-class smoothness is improved, which guarantee the robustness of the model.

\subsection{Analysis on Margin $\bm{m}$}
\label{sec:m}
We use MNIST~\cite{LeCun98} to conduct adversarial study~\cite{Szegedy14,Goodfellow15,Na18,Yan18}, to further analyze our MTER method. The only hyperparameter in our method is the margin $m$. So we would like to explore the influence of $m$ and give advice on choice of it under different settings.

\begin{figure}[htbp]
	\center
	\includegraphics[width=0.95\linewidth]{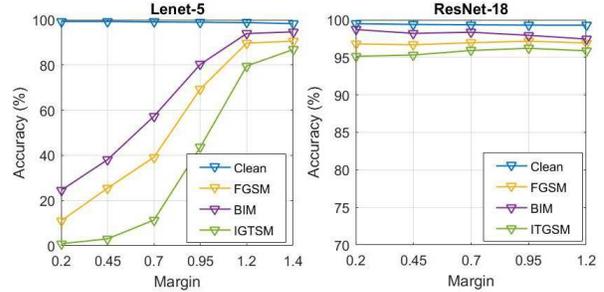}
	\caption{LeNet-5~\cite{lecun-98} and ResNet-18~\cite{he2016deep} trained using MTER by varying margin $m$ in $\{0.2,0.45,0.7,0.95,1.2,1.4\}$. The two type of models both could obtain good test accuracy on MNIST~\cite{LeCun98}. However, for LeNet-5~\cite{lecun-98}, along with the increase of margin $m$, the robustness to attacks improves gradually. While for ResNet-18~\cite{he2016deep}, the robustness has reached a relative high level and remains unchanged when the margin increases.}
	\vspace{-1.5ex}
	\label{fig:margin}
\end{figure}

First we train LeNet-5~\cite{lecun-98} and ResNet-18~\cite{he2016deep} by varying $m$ in $\{0.2,0.45,0.7,0.95,1.2,1.4\}$. Then we test these models using aforementioned attack methods, \ie FGSM~\cite{Goodfellow15}, BIM~\cite{Kurakin17} and ITGSM~\cite{Kurakin17}. The results are illustrated in Figure~\ref{fig:margin}, from which we can discover significant difference between the two type of models, LeNet-5~\cite{lecun-98} and ResNet-18~\cite{he2016deep}. Although the two type of models both could obtain good test accuracy on MNIST~\cite{LeCun98}, the accuracy for LeNet-5~\cite{lecun-98} and ResNet-18~\cite{he2016deep} is 99.2\% and 99.6\% respectively. However, for LeNet-5~\cite{lecun-98}, along with the increase of margin $m$, the robustness to different attacks improves gradually and the accuracy on clean images decreases slightly. While for ResNet-18~\cite{he2016deep}, both the accuracy on clean test set and the robustness against adversarial attacks, have reached a relative high level and remain unchanged when the margin increases. 

\begin{figure}[htbp]
	\center
	\includegraphics[width=0.95\linewidth]{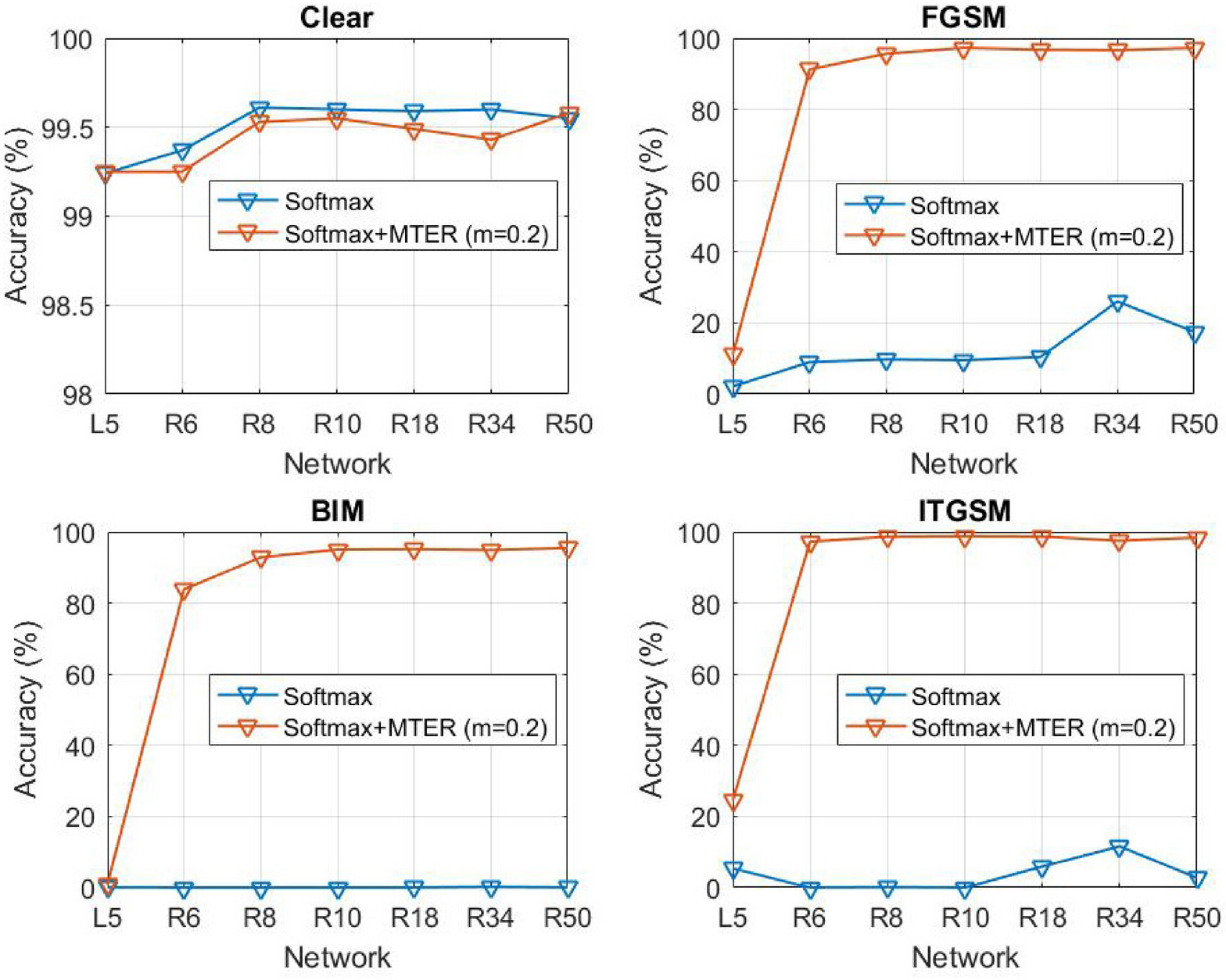}
	\caption{Accuracy on clean images, and adversarial examples generated by FGSM~\cite{Goodfellow15}, BIM~\cite{Kurakin17}, and ITGSM~\cite{Kurakin17} of models with different architectures. ``L5'',``R6'',``R8'',``R10'',``R18'',``34'' and ``R50'' on the x-axis denote LeNet-5~\cite{lecun-98}, ResNet-6~\cite{he2016deep}, ResNet-8~\cite{he2016deep}, ResNet-10~\cite{he2016deep}, ResNet-18~\cite{he2016deep}, ResNet-34~\cite{he2016deep} and ResNet-50~\cite{he2016deep}. For models trained using Softmax combined with MTER($m=0.2$), the accuracy on the clean images and three adversarial examples increases when the model size becomes bigger (from LeNet-5~\cite{lecun-98} to ResNet-10~\cite{he2016deep}), while remain relatively stable when the model size reaches a certain value (after ResNet-10~\cite{he2016deep})}
	\vspace{-1.5ex}
	\label{fig:network}
\end{figure}

Furthermore, we fix the margin $m=0.2$. We use this relative small margin because we would like to observe the performance of different networks under relaxed state of our method. We train models under Softmax and our MTER method respectively, using different architectures, \eg LeNet-5~\cite{lecun-98}, ResNet-6~\cite{he2016deep}, ResNet-8~\cite{he2016deep}, ResNet-10~\cite{he2016deep}, ResNet-18~\cite{he2016deep}, ResNet-34~\cite{he2016deep} and ResNet-50~\cite{he2016deep}. We still test these trained models under the three aforementioned attacks. The results are shown in Figure~\ref{fig:network}. In the figure~\ref{fig:network}, for models trained using Softmax, the accuracy on the clean images increases when the model size becomes bigger (from LeNet-5~\cite{lecun-98} to ResNet-50~\cite{he2016deep}). For models trained using Softmax combined with MTER ($m=0.2$), the accuracy on the clean images and three adversarial examples increases when the model size becomes bigger (from LeNet-5~\cite{lecun-98} to ResNet-10~\cite{he2016deep}), while remain relatively stable when the model size reaches a certain value (after ResNet-10~\cite{he2016deep}). 

We then infer that a specific classification task need a certain amount of computing power to fit the clean set and its augmentation set, \eg adversarial examples. It is easy for our method to push a large model to learn both the clean images and the adversarial examples under a relative relaxed state (small $m$), while it will not work for a relative small model with less computing power. Therefore we recommend to increase the margin $m$ to push the ``lazy'' model fighting with adversarial examples but sacrificing a little performance on the original dataset,  if a small model is used and is still concerned about robustness.  

\subsection{Black-box Attack Analysis}
We also use MNIST~\cite{LeCun98} for analysis of adversarial attacks and defense under black-box settings. We report black box attack accuracy of the adversarial examples generated from a source network and tested on another target network. In our experiment, both the source network and the target network are trained with adversarial learning methods in different architectures. Specifically, we use four models, the architecture of them are LeNet-5~\cite{lecun-98} or ResNet-18~\cite{he2016deep}, and training methods are adversarial training~\cite{Kurakin17atscale} or our MTER method. The adversarial examples from the source models are generated by FGSM~\cite{Goodfellow15} or BIM~\cite{Kurakin17} with $\varepsilon=0.3*255$.

The results on FGSM~\cite{Goodfellow15} and BIM~\cite{Kurakin17} are shown in Table~\ref{table:black_fgsm} and Table~\ref{table:black_bim}, respectively. The row and the column denote the source and target model respectively. The LeNet-5~\cite{lecun-98} is denoted as ''L5``, and ResNet-18~\cite{he2016deep} is denoted as ''R18``. ''ADV`` is a shorthand of adversarial training~\cite{Kurakin17atscale}. ''MTER-R18`` means this ResNet-18~\cite{he2016deep} model is trained supervised by Softmax combined with our MTER method. The higher the accuracy is, the more robust is the target model. 

As shown in the Table~\ref{table:black_fgsm} and Table~\ref{table:black_bim}, if the source models and the architecture of targets models are both equal, the accuracy of target models trained using MTER method is higher than that of adversarial training method~\cite{Kurakin17atscale}. This phenomenon indicates our MTER method show better robust performance under black box attack scenario, on both FGSM~\cite{Goodfellow15} and BIM~\cite{Kurakin17} adversarial examples, even if the adversarial training have used FGSM~\cite{Goodfellow15} examples for augmentation. Besides, we find that the ResNet-18~\cite{he2016deep} models are more robust than LeNet-5~\cite{lecun-98} models, while adversarial examples generated from LeNet-5~\cite{lecun-98} are more aggressive and have better transferability than those of ResNet-18~\cite{he2016deep}. 

\begin{table}
	\renewcommand\arraystretch{1}
	\begin{center}
		\scalebox{0.75}{
			\begin{tabular}{|c||c|c||c|c|}
			\hline
			\diagbox{Source}{Target}&ADV-L5&MTER-L5&ADV-R18&MTER-R18\\
			\hline\hline
			ADV-L5&\diagbox{}{}&92.0&95.6&97.3\\
			MTER-L5&85.3&\diagbox{}{}&94.5&97.7\\
			ADV-R18&89.8&94.9&\diagbox{}{}&97.4\\
			MTER-R18&45.2&94.2&94.1&\diagbox{}{}\\
			\hline
			\end{tabular}
		}
	\end{center}
	\vspace{-0.2cm}
	\caption{MNIST~\cite{LeCun98} test result (\%) on FGSM~\cite{Goodfellow15} ($\varepsilon=0.3*255$) adversarial examples under black box settings. The row and the column denote the source and target model respectively. The LeNet-5~\cite{lecun-98} is denoted as ''L5``, and ResNet-18~\cite{he2016deep} is denoted as ''R18``. ''ADV`` is a shorthand of adversarial training~\cite{Kurakin17atscale}. The higher the accuracy is, the more robust is the target model.}
\vspace{-1.5ex}
	\label{table:black_fgsm}
\end{table}

\begin{table}
	\renewcommand\arraystretch{1}
	\begin{center}
		\scalebox{0.75}{
			\begin{tabular}{|c||c|c||c|c|}
				\hline
				\diagbox{Source}{Target}&ADV-L5&MTER-L5&ADV-R18&MTER-R18\\
				\hline\hline
				ADV-L5&\diagbox{}{}&90.1&89.1&97.3\\
				MTER-L5&79.3&\diagbox{}{}&92.8&97.5\\
				ADV-R18&88.2&94.9&\diagbox{}{}&97.3\\
				MTER-R18&84.0&94.6&95.6&\diagbox{}{}\\
				\hline
			\end{tabular}
		}
	\end{center}
\vspace{-0.2cm}
	\caption{MNIST~\cite{LeCun98} test result (\%) on BIM~\cite{Kurakin17} ($\varepsilon=0.3*255$) adversarial examples under black box settings. The row and the column denote the source and target model respectively. The LeNet-5~\cite{lecun-98} is denoted as ''L5``, and ResNet-18~\cite{he2016deep} is denoted as ''R18``. ''ADV`` is a shorthand of adversarial training~\cite{Kurakin17atscale}. The higher the accuracy is, the more robust is the target model.}
	\vspace{-1.5ex}
	\label{table:black_bim}
\end{table}

\subsection{Experiment on Deep Face Recognition}
\label{section:face}
In a deep face recognition system, an adversary may try to disguise a face as an authorized user. We simulate this scenario using state-of-art face recognition models and test our MTER method. Deep face recognition is a open-set problem, which indicates the training identities and the test identities are usually different. We don't directly classify an identity by end-to-end classification probability, but use a DNN as a deep feature extracter and compare deep features to distinguish faces.

\textbf{Training datasets.} In the experiment, the training datasets are CASIA-WebFace~\cite{Yi2014CASIA}, VGGFace2~\cite{Cao18} and MS1M-IBUG~\cite{Guo16MS}. The CASIA-WebFace~\cite{Yi2014CASIA} dataset is the first widely used large-scale training dataset in deep face recognition, containing 0.49M images from 10,575 celebrities. VGGFace2~\cite{Cao18} is a large-scale dataset containing 3.31M images from 9131 celebrities. There are diverse and abundant images in VGGFace2~\cite{Cao18}, which have large variations in pose, age, illumination, ethnicity and profession. MS1M-IBUG~\cite{Guo16MS} (referred to as MS1M~\cite{Guo16MS}) is a refined version of MS-Celeb-1M dataset~\cite{Guo16MS}, which is public available and widely used. The original MS-Celeb-1M dataset~\cite{Guo16MS} contains about 10k celebrities with 10M images. MS1M~\cite{Guo16MS} is refined by Deng \etal~\cite{deng2018arcface} to decrease the noise and finally contains 3.8M images of 85,164 celebrities. 

\textbf{Network settings.} For all the embedding networks, we adopt the ResNet-50~\cite{he2016deep}, but make changes as~\cite{deng2018arcface} which apply the ``BN~\cite{ioffe2015batch}-Dropout~\cite{Srivastava2014Dropout}-FC-BN'' sturcture to get the final 512-$D$ embedding feature. For image preprocessing, the images are cropped and aligned to the normalized $112\times112$ face following~\cite{deng2018arcface}. In the training process, the original models are supervised by an effective loss function ArcFace~\cite{deng2018arcface}, which has been widely accepted by industry. The feature scale is set to 64 and angular margin is set to 0.5 for CASIA-WebFace~\cite{Yi2014CASIA} and MS1M~\cite{Guo16MS}, and 0.3 for VGGFace2~\cite{Cao18} following the original paper~\cite{deng2018arcface}. Different from the object classification experiment, the face models are finetuned with ArcFace combined with our MTER method. The face models finetuned with MTER method have fine convergence speed and usually converge in no more than 3 epoches. We set $m$ to 0.2, 1.2 and 1.4 for CASIA-WebFace~\cite{Yi2014CASIA}, VGGFace2~\cite{Cao18} and MS1M~\cite{Guo16MS}, respectively.

\begin{figure}[htbp]
	\center
	\includegraphics[width=0.95\linewidth]{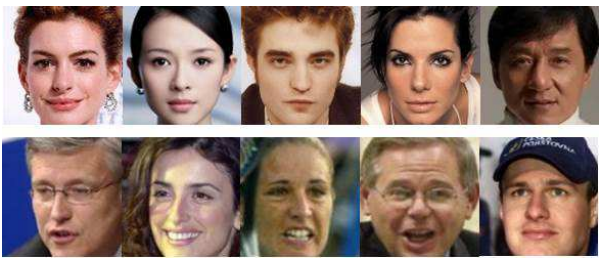}
	\caption{The first row is the five target identities, the second row is five attackers which are randomly selected in all the 13233 attackers.}
	\vspace{-1.5ex}
	\label{fig:face}
\end{figure}

\begin{figure}[htbp]
	\center
	\includegraphics[width=0.95\linewidth]{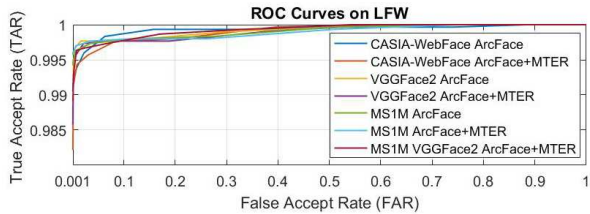}
	\caption{ROC curves of different models on LFW~\cite{LFWTech}. We define the distance threshold of a network for attacking (or distinguishing a positive and a negative pair) to have a low false accept rate (FAR = $1e-3$) on LFW.}
	\vspace{-1.5ex}
	\label{fig:ROC}
\end{figure}

\begin{table*}
	\renewcommand\arraystretch{1}
	\begin{center}
		\scalebox{0.75}{
			\begin{tabular}{|c|c||c|c|c||c|c|c|}
				\hline
				Training Method&Training Set&FFTGSM($\varepsilon$=10)&IFTGSM($\varepsilon$=5)&IFTGSM($\varepsilon$=10)&ITGSM~\cite{Kurakin17}($\varepsilon$=10)&LFW&YTF\\
				\hline\hline
				ArcFace~\cite{deng2018arcface}&CASIA-WebFace~\cite{Yi2014CASIA}&99.3&100.0&100.0&98.3&99.5&95.6\\
				ArcFace~\cite{deng2018arcface}+adv.&CASIA-WebFace~\cite{Yi2014CASIA}&5.1 ($\downarrow$ 94.2)&5.7 ($\downarrow$ 94.3)&49.5 ($\downarrow$ 50.5)&0.8 ($\downarrow$ 97.5)&99.4&94.7\\
				ArcFace~\cite{deng2018arcface}+MTER&CASIA-WebFace~\cite{Yi2014CASIA}&2.0 ($\downarrow$ 97.3)&3.5 ($\downarrow$ 96.5)&27.4 ($\downarrow$ 72.6)&0.1 ($\downarrow$ 98.2)&99.5&94.8\\
				\hline\hline
				ArcFace~\cite{deng2018arcface}&VGGFace2~\cite{Cao18}&98.3&100.0&100.0&100.0&99.7&97.7\\
				ArcFace~\cite{deng2018arcface}+adv.&VGGFace2~\cite{Cao18}&3.1 ($\downarrow$ 95.2)&5.5 ($\downarrow$ 94.5)&63.8 ($\downarrow$ 36.2)&0.1 ($\downarrow$ 99.9)&99.5&97.2\\
				ArcFace~\cite{deng2018arcface}+MTER&VGGFace2~\cite{Cao18}&4.6 ($\downarrow$ 93.7)&6.1 ($\downarrow$ 93.9)&35.6 ($\downarrow$ 64.4)&0.1 ($\downarrow$ 99.9)&99.6&97.5\\
				\hline\hline
				ArcFace~\cite{deng2018arcface}&MS1M~\cite{Guo16MS}&99.8&100.0&100.0&69.2&99.7&97.0\\
				ArcFace~\cite{deng2018arcface}+adv.&MS1M~\cite{Guo16MS}&45.4 ($\downarrow$ 54.4)&20.1 ($\downarrow$ 79.9)&62.6 ($\downarrow$ 37.4)&0.1 ($\downarrow$ 	69.1)&99.6&96.2\\
				ArcFace~\cite{deng2018arcface}+MTER&MS1M~\cite{Guo16MS}&4.1 ($\downarrow$ 95.7)&7.9 ($\downarrow$ 92.1)&61.4 ($\downarrow$ 38.6)&0.0 ($\downarrow$ 69.2)&99.8&96.9\\
				ArcFace~\cite{deng2018arcface}+MTER&MS1M~\cite{Guo16MS}+VGGFace2~\cite{Cao18}&9.6 ($\downarrow$ 90.2)&6.2 ($\downarrow$ 93.8)&19.5 ($\downarrow$ 80.5)&0.1 ($\downarrow$ 69.1)&99.5&96.8\\
				\hline
			\end{tabular}
		}
	\end{center}
\vspace{-0.2cm}
	\caption{The average hit rate of models trained on CASIA-WebFace~\cite{Yi2014CASIA}, VGGFace2~\cite{Cao18} and MS1M~\cite{Guo16MS} supervised by ArcFace loss~\cite{deng2018arcface}, ArcFace~\cite{deng2018arcface}+adv., and ArcFace~\cite{deng2018arcface}+MTER, respectively. Attacks are launched from attackers to disguise targets in the feature level using FFTGSM, IFTGSM, and in the label level using ITGSM~\cite{Kurakin17}. The lower is the hit rate, the stronger is robustness of models.}
	\label{table:face_robust_p}
	\vspace{-1.5ex}
\end{table*}

\textbf{Recognition performance.} We test the recognition performance of all the models on LFW~\cite{LFWTech} and YTF~\cite{Wolf2011Face}. LFW~\cite{LFWTech} contains 13233 face images from 5749 different identities. We follow the unrestricted with labeled outside data protocol on LFW~\cite{LFWTech} and test on the 3000 positive (same identity) and 3000 (different identity) negative pairs. YTF~\cite{Wolf2011Face} is a database of face video collected from YouTube, which consists of 3,425 videos of 1,595 different people. Each video varies from 48 to 6,070 frames, with an average length as 181.3 frames. We follow the unrestricted with labeled outside data protocol on all the test datasets.  

\textbf{Robustness performance.} To simulate the face disguise scenario, we select five person as target identities, as shown in the first row of Figure~\ref{fig:face}. Then we use the 13233 face images in LFW~\cite{LFWTech} as attackers to disguise another five target person respectively, which construct a $13233\times5$ attack matrix to simulate random attacks. We test the robustness under two attack settings:(1) feature attacks and (2) label attacks. The feature attack is more practical in face recognition, while we use the label attack for demonstrating the effectiveness of our method in label defense of close-set settings which is a rarity in deep face recognition. 

First we define the feature attacks settings. The attacks are launched from attackers to disguise targets. Specifically, the attack goal is to get face embedding representations of attackers closer to those of targets than the distance threshold of a face recognition system. Next, we define the threshold of a DNN in our simulation. Using positive and negative pairs of LFW~\cite{LFWTech}, we compute the Euclidean distance of normalized deep features to get ROC curves, as shown in Figure~\ref{fig:ROC}. Then we identify distance thresholds for judging a pair is positive or negative. Since we would like to compare the adversarial robustness of the trained models like real-world applications, we define the distance threshold for attacking (or distinguishing a positive and a negative pair) to have a low false accept rate (FAR = $1e-3$). We will generate attacks~\eqref{a:3} using IFTGSM~\eqref{a:5} and FFTGSM~\eqref{a:7}. Then we define the evaluation criteria to measure the robustness of models.  The attack goal is to get face embedding representations of attackers closer to those of targets than the Euclidean distance threshold.  The defense goal is to keep the distance between the representations of attackers and targets larger than the threshold. Therefore, an attack is defined as a hit if the embedding distance between the attacker and target is lower than the threshold. We use the average hit rate of the five targets to report the robustness performance of the trained models. The lower is the average hit rate, the stronger is robustness of the model. 

Finally we introduce the label attacks settings. Although the training identities are different from the test ones, we could use the predicted identity of the targets as their labels and let the attackers to launch attacks towards the predicted labels. Meanwhile, a hit is defined as the predicted label of an attacker is the same as that of the target. We also use the average hit rate of the five targets to report the robustness performance. ITGSM~\cite{Kurakin17} will be used to generate label attacks. We will not report result of FTGSM~\cite{Kurakin17} attacks because we find the this method often fails to attack face models.  

\textbf{Results.} The results of defense performance against feature and label level attacks are listed in Table~\ref{table:face_robust_p}. The hit rates of original models are close to 100 percent under settings where 13233 different attackers disguise targets. This may indicate that the state-of-art face models are indeed highly vulnerable to adversarial attacks, and an arbitrary attackers would have high probability to disguise another identity. While with our MTER method, the hit rate decrease significantly, which indicates that our method improve the robustness of the state-of-art face models in both open-set and close-set settings and prevent the face disguise feature attacks to a certain degree. Besides, We discover that the robust performance of our method on MS1M~\cite{Guo16MS} is less significant than that on CASIA-WebFace~\cite{Yi2014CASIA} and VGGFace2~\cite{Cao18}. Therefore we recommend to use the large datasets with less identities to finetune the models with large identities to get better robustness performance, \eg finetune the original model trained on MS1M~\cite{Guo16MS} using VGGFace2~\cite{Cao18}. To further evaluate our method, we also compare with a strong baseline by finetuning the original models and incorporating adversarial examples generated using IFTGSM~\eqref{a:5}. The result shows that our method further benefits from additional embedding regularization, which indicates that incorporating adversarial examples in the training process could improve robustness, while how to optimize with them is also crucial. 

The results of face recognition performance on original models and robust models are shown in Table~\ref{table:face_r_p}. We also list the state-of-art models in face recognition community. We could also observe that, the accuracy of robust models on LFW~\cite{LFWTech} and YTF~\cite{Wolf2011Face} decreased slightly, which indicates that we may sacrifice a certain degree of recognition performance for the improvement of adversarial robustness. 

\begin{table}
	\renewcommand\arraystretch{1}
	\begin{center}
		\scalebox{0.74}{
			\begin{tabular}{|c|c|c|c|}
				\hline
				Training Method&Training Set&LFW&YTF\\
				\hline\hline
				DeepFace~\cite{6909616}&4M&97.35&91.4\\
				FaceNet~\cite{Schroff2015FaceNet}&200M&99.63&95.1\\
				VGG Face~\cite{parkhi2015deep}&2.6M&98.95&97.3\\
				DeepID2+~\cite{sun2015deeply}&0.3M&99.47&93.2\\
				Center Face~\cite{Wen2016A}&0.7M&99.28&94.9\\
				Noisy Softmax~\cite{chen2017noisy}&WebFace+&99.18&94.88\\
				Triplet Loss~\cite{Schroff2015FaceNet}&WebFace~\cite{Yi2014CASIA}&98.70&93.4\\
				L-Softmax Loss~\cite{Liu2016Large}&WebFace~\cite{Yi2014CASIA}&99.10&94.0\\
				Softmax+Center Loss~\cite{Wen2016A}&WebFace~\cite{Yi2014CASIA}&99.05&94.4\\
				SphereFace~\cite{Liu2017SphereFace}&WebFace~\cite{Yi2014CASIA}&99.42&95.0\\
				CosFace~\cite{Wang2018CosFace}&WebFace~\cite{Yi2014CASIA}&99.33&96.1\\
				ArcFace~\cite{deng2018arcface}&MS1MV2 (5.8M)&99.83&98.02\\
				\hline\hline
				ArcFace~\cite{deng2018arcface}&WebFace~\cite{Yi2014CASIA}&99.5&95.6\\
				ArcFace~\cite{deng2018arcface}+adv.&WebFace~\cite{Yi2014CASIA}&99.4&94.7\\
				ArcFace~\cite{deng2018arcface}+MTER&WebFace~\cite{Yi2014CASIA}&99.5&94.8\\
				\hline\hline
				ArcFace~\cite{deng2018arcface}&VGGFace2~\cite{Cao18}&99.7&97.7\\
				ArcFace~\cite{deng2018arcface}+adv.&VGGFace2~\cite{Cao18}&99.5&97.2\\
				ArcFace~\cite{deng2018arcface}+MTER&VGGFace2~\cite{Cao18}&99.6&97.5\\
				\hline\hline
				ArcFace~\cite{deng2018arcface}&MS1M~\cite{Guo16MS}&99.7&97.0\\
				ArcFace~\cite{deng2018arcface}+adv.&MS1M~\cite{Guo16MS}&99.6&96.2\\
				ArcFace~\cite{deng2018arcface}+MTER&MS1M~\cite{Guo16MS}&99.8&96.9\\
				ArcFace~\cite{deng2018arcface}+MTER&MS1M~\cite{Guo16MS}+VGGFace2~\cite{Cao18}&99.5&96.8\\
				\hline
			\end{tabular}
		}
	\end{center}
	\vspace{-0.2cm}
	\caption{The accuracy on LFW~\cite{LFWTech} and YTF~\cite{Wolf2011Face}. The state-of-art models in face recognition community are listed in the first cell. Other cells are our models used in the face disguise experiment.}
	\vspace{-1.5ex}
	\label{table:face_r_p}
\end{table}

\section{Conclusion}
We have proposed a margin-based triplet embedding regularization (MTER) method to improve the robustness of DNNs. Experiments on MNIST~\cite{LeCun98}, CASIA-WebFace~\cite{Yi2014CASIA}, VGGFace2~\cite{Cao18} and MS1M~\cite{Guo16MS} have demonstrated the effectiveness of our method in simple object classification and deep face recognition. 

\section{Acknowledgments}
This work was partially supported by the National Natural Science Foundation of China under Grant Nos. 61573068 and 61871052.

{\small
\bibliographystyle{ieee}
\bibliography{4562_final}
}

\end{document}